\documentclass[10pt,twocolumn,letterpaper]{article}

\usepackage{iccv}
\usepackage{times}
\usepackage{epsfig}
\usepackage{graphicx}
\usepackage{amsmath}
\usepackage{amssymb}
\usepackage{multirow}

\usepackage[breaklinks=true,bookmarks=false]{hyperref}

\iccvfinalcopy % *** Uncomment this line for the final submission

\def\iccvPaperID{****} % *** Enter the ICCV Paper ID here

\ificcvfinal\fi

\begin{document}

%%%%%%%%% TITLE
\title{Coarse to Fine: Video Retrieval before Moment Localization}

% \author{Zijian Gao\\
% BUPT, Tencent\\
% {\tt\small gzjbupt2016@bupt.edu.cn}
% \and
% Huanyu Liu\\
% TJU\\
% Tencent\\
% {\tt\small asdfjkhqwe@asdfjqw.cn}
% \and
% Jingyu Liu\\
% Tencent\\
\author{Zijian Gao\textsuperscript{1,3}, Huanyu Liu\textsuperscript{2,3}, Jingyu Liu\textsuperscript{3}\\
\textsuperscript{1}Beijing University of Posts and Telecommunications\\
\textsuperscript{2}Tianjin University\\
\textsuperscript{3}Tencent PCG
}

\maketitle
\ificcvfinal\thispagestyle{empty}\fi

%%%%%%%%% ABSTRACT
\begin{abstract}
The current state-of-the-art methods for video corpus moment retrieval (VCMR) often use similarity-based feature alignment approach for the sake of convenience and speed. 
However, late fusion methods like cosine similarity alignment are unable to make full use of the information from both query texts and videos. 
In this paper, we combine feature alignment with feature fusion to promote the performance on VCMR.
\end{abstract}

%%%%%%%%% BODY TEXT
\section{Introduction}

Video Corpus Moment Retrieval (VCMR) \cite{escorcia2019temporal} is a new video-text retrieval task which aims to retrieve the most relevant moments from a large video corpus instead of from a single video. 
The text-based VCMR can be decomposed into two sub-tasks: video retrieval (VR) and single video moment retrieval (SVMR). 
The former requires to retrieve the most relevant video and the goal of the latter is retrieval the most relevant moment from a single video. 

We explore that the methods proposed recently \cite{li2020hero, zhang2021video} mainly following \cite{lei2020tvr} which only aligns video feature and query feature by calculating similarity. 
Different from that, methods fusing video with query \cite{zhang2020span, zhang2021parallel} are popular across SVMR realm, since they can align video and query within a more fine-grained scope and collect adequate information. 
Therefore, in this paper, we combine conventional VCMR method with SVMR method to retrieve video moments more precisely.

%------------------------------------------------------------------------
\section{Approach}

\subsection{Reminiscence of HERO}

To facilitate the description, we use the same mathematical symbols as in \cite{li2020hero}.
Given a video denoted by series of frames $\mathbf{v} =  \{v_i\}^{N_v}_{i=1}$ and subtitles $\mathbf{s} = \{s_i\}^{N_s}_{i=1}$, where $N_v$ is the number of frames in the video and $N_s$ is the number of sentences in its subtitles. Frame embeddings $\mathbf{V}^{emb}_{s_i} \in \mathbb{R}^{K \times D}$ and token embeddings $\mathbf{W}^{emb}_{s_i} \in \mathbb{R}^{L \times D}$ are encoded by frame embedder and text embedder, respectively.

The alignment between tokens and frames helps model collect abundant information.
Therefore, a Cross-Modal Transformer with cross-modal attention for multimodal fusion is adopted to get a sequence of contextualized embeddings:
\begin{equation}
    \mathbf{V}^{cross}_{s_i}, \mathbf{W}^{cross}_{s_i} = \mathit{f}_{cross}(\mathbf{V}^{emb}_{s_i}, \mathbf{W}^{emb}_{s_i}),    
\end{equation}
where $\mathit{f}_{cross}(\cdot, \cdot)$ denotes the cross-modal transformer.

In order to make full use of temporal information, contextualized frame embeddings are re-organized into a video $ \mathbf{V}^{cross}=\{\mathbf{V}^{cross}_{s_i}\}^{N_s}_{i=1} \in \mathbb{R}^{N_v \times d} $ following frame order, and a Temporal Transformer is utilized to learn global context of a video. The final contextualized video embeddings are calculated as :
\begin{equation}
    \mathbf{V}^{temp} = \mathit{f}_{temp}(\mathbf{V}^{emb}+\mathbf{V}^{cross}),    
\end{equation}
where $\mathit{f}_{temp}(\cdot)$ is the temporal transformer.

As for query, the query text is fed into cross-modal transformer to compute query embeddings $\mathbf{W}^{cross}_{s_q}=\mathit{f}_{cross}(\mathbf{0}, \mathbf{W}^{emb}_{s_q}) \in \mathbb{R}^{N_q \times d}$.
Then a query encoder takes query embedings as input and outputs the final query vector $\mathbf{q} \in \mathbb{R}^{d}$.

The similarity score of the whole video and the query text is computed by max-pooling the cosine similarity between each frame and the query text:
\begin{equation}
    S_{global}(s_q,\mathbf{v})=\max\left(\frac{\mathbf{V}^{temp}}{\|\mathbf{V}^{temp}\|}\frac{\mathbf{q}}{\|\mathbf{q}\|}\right).    \label{con:global similarity}
\end{equation}

Local similarity used for moment retrieval is computed by dot product:
\begin{equation}
    S_{local}(s_q,\mathbf{v})=\mathbf{V}^{temp}\mathbf{q},    
\end{equation}
and two trainable 1D convolution filters are applied to local similarity score to acquire $\mathbf{p}_{st}$ and $\mathbf{p}_{ed}$ which are the probabilities of each frame being the start or end frame of the moment.

\subsection{Moment Retrieval with fusion}

As mentioned above, the moment retrieval of HERO is a late fusion method based on similarity score. 
However, retrieving moment by using local similarity neglect the abundant information between two modalities. 
To enhance cross-modal interactions, we utilize context-query attention (CQA) \cite{yu2018qanet, zhang2020span} to integrate video and query. 
The similarity $\mathbf{S} \in \mathbb{R}^{N_v \times N_q}$ between each frame and each query token is calculated. 
Then the context-to-query attention $\mathbf{A}$ and query-to-context attention $\mathbf{B}$ are calculated by:
\begin{equation}
    \mathbf{A}=\mathbf{S}_r\cdot\mathbf{W}^{emb}_{s_q},\mathbf{B}=\mathbf{S}_r\cdot\mathbf{S}^T_c\cdot\mathbf{V}^{temp},    
\end{equation}
where $\mathbf{S}_r$ and $\mathbf{S}_c$ is the row- and column-normalized $\mathbf{S}$ by $\mathit{softmax}$, respectively. 
And the final feature of fusion is the concatenation of original features and attention guided features that:
\begin{equation}
    \mathbf{V}^q = \mathit{FC}([\mathbf{V}^{temp};\mathbf{A};\mathbf{V}^{temp}\odot\mathbf{A};\mathbf{V}^{temp}\odot\mathbf{B}]),    
\end{equation}
where $\mathit{FC}$ is a fully-connected layer reducing feature dimensions to $d$; $\odot$ means element-wise multiplication.

We split a video into two parts with the target moment as foreground and the rest as background. 
To further distinguish foreground and background, our model needs to learn the importance of each frame with the whole query text. 
To this end, an another query encoder is applied for query embeddings $\mathbf{W}^{emb}_{s_q}$ to get the whole sentence feature $\mathbf{q}_c$. 
$\mathbf{q}_c$ is concatenated with frame embeddings as $\bar{\mathbf{V}}^q=\{\bar{\mathbf{V}}^q_{i}\}^{N_v}_{i=1}$, where $\bar{\mathbf{V}}^q_i=[\mathbf{V}^q_i;\mathbf{q}_c]$. 
And the importance score is calculated by:
\begin{equation}
    \mathbf{S}_h = \mathit{sigmoid}(\mathit{Conv1D}(\bar{\mathbf{V}}^q)) \in \mathbb{R}^{N_v},    
\end{equation}
Then the importance score can highlight the fused features:
\begin{equation}
    \tilde{\mathbf{V}}^q = \mathbf{S}_h\cdot\mathbf{V}^q.    
\end{equation}

According to \cite{zhang2020span}, two-layer LSTMs with two fully-connected layers serve as the conditioned span predictor to generate the scores:

\begin{equation}
    \begin{aligned}
        \mathbf{h}^{st}_t &= \mathit{LSTM}_{st}(\tilde{\mathbf{V}}^q, \mathbf{h}^{st}_{t-1}),\\
        \mathbf{h}^{ed}_t &= \mathit{LSTM}_{ed}(\mathbf{h}^{st}_t, \mathbf{h}^{ed}_{t-1}),\\
        \mathbf{S}^{st}_t &= \mathit{FC}_{st}([\mathbf{h}^{st}_t;\tilde{\mathbf{V}}^q_t]),\\
        \mathbf{S}^{ed}_t &= \mathit{FC}_{ed}([\mathbf{h}^{ed}_t;\tilde{\mathbf{V}}^q_t]),
    \end{aligned}
\end{equation}

where $\mathit{FC}$ is the fully-connected layer; $\mathbf{S}^{st}_t$ and $\mathbf{S}^{ed}_t$ denote the scores of every position being the start and end of the ground-truth span.

The training objective of moment retrieval is defined as:
\begin{equation}
\begin{aligned}
    \mathcal{L} &= 0.5 * (\mathcal{L}_{ce}(\mathbf{S}^{st}, Y_{st}) + \mathcal{L}_{ce}(\mathbf{S}^{ed}, Y_{ed})) \\
                &+ \mathcal{L}_{l1}(\mathbf{S_h, Y_h})
\end{aligned}
\end{equation}
where $\mathcal{L}_{ce}$ and $\mathcal{L}_{l1}$ are cross-entropy loss and L1 loss.

Note that only positive video-query pairs are fed into moment retrieval training. 
In inference, we perform two stage operation that similarity scores between each query text and each video are computed by Eq.\eqref{con:global similarity} quickly, then top-$k$ ranked videos are selected for each query text to localize moments through fusion method.

\subsection{Pre-training}

Since retrieval task depends on the quality of visual representations, 
ViT-CLIP+S3D features are selected as our inputs of visual modality. 
However, this may damage the performance of the provided pre-trained model which is trained with ResNet+S3D. 
Therefore, following the pre-training strategy used in HERO, we re-train our model with four pre-training tasks.

\subsection{Data Augmentation}

To promote the robustness, some data augmentations are used in training.
Following \cite{yan2021consert}, we adopt token shuffling, cutoff and dropout.
Token shuffling strategy aims to randomly shuffle order of the tokens in the token embeddings.
Cutoff consists of token cutoff and feature cutoff which aims to randomly erase some tokens or feature dimension.
Dropout is a widely used method to mitigate over-fitting. We utilize it to randomly drop some elements in the token embeddings and set values to zero.
The data augmentation operations are imposed on video embeddings, subtitle token embeddings and query token embeddings with a probability of $50\%$.
We decompose the $50\%$ into $40\%$ token shuffling, $15\%$ token cutoff, $15\%$ feature cutoff, $15\%$ dropout and $15\%$ unchanged.

\section{Experiments}

\subsection{Dataset and Evaluation Metric}

This paper is a technical report about VALUE Challenge and we only focu on retrieval task. 
We conduct experiments on two VCMR datasets, TVR and How2R, And two VR datasets, YC2R and VATEX-EN-R.
According to VALUE benchmark \cite{li2021value}, either VCMR task or VR task uses Average recall at K (R@K) as evaluation metrics, and Average Recall (AveR) at \{1, 5, 10\} is also adopted for overall appraisal.

\subsection{Performance Comparison on VCMR}

\begin{table}[htbp]
    \footnotesize
    \begin{center}
    \setlength{\tabcolsep}{1.5mm}{
    \begin{tabular}{c|c|ccc|ccc}
         \hline
         
         \multicolumn{1}{c|}{}&\multicolumn{1}{c|}{\multirow{2}{*}{Method}} & \multicolumn{3}{c|}{TVR} &\multicolumn{3}{c}{How2R}\\
         
         & & {\scriptsize R@1} & {\scriptsize R@10} & {\scriptsize R@100} & {\scriptsize R@1} & {\scriptsize R@10} & {\scriptsize R@100}\\
         \hline
         \hline
        \multicolumn{1}{c|}{\multirow{6}{*}{\rotatebox[origin=c]{90}{VAL}}} & \multicolumn{1}{c|}{XML \cite{lei2020tvr}}       & \multicolumn{1}{c}{2.62}    & \multicolumn{1}{c}{9.05}     & \multicolumn{1}{c|}{22.47}        & \multicolumn{1}{c}{-}    & \multicolumn{1}{c}{-}     & \multicolumn{1}{c}{-}       \\
        & \multicolumn{1}{c|}{HERO \cite{li2020hero}}       & \multicolumn{1}{c}{5.13}    & \multicolumn{1}{c}{16.26}     & \multicolumn{1}{c|}{24.55}      & \multicolumn{1}{c}{-}    & \multicolumn{1}{c}{-}     & \multicolumn{1}{c}{-}           \\
        & \multicolumn{1}{c|}{HAMMER \cite{zhang2020hierarchical}}       & \multicolumn{1}{c}{5.13}    & \multicolumn{1}{c}{11.38}     & \multicolumn{1}{c|}{16.71}     & \multicolumn{1}{c}{-}    & \multicolumn{1}{c}{-}     & \multicolumn{1}{c}{-}       \\
        & \multicolumn{1}{c|}{ReLoCLNet \cite{zhang2021video}}       & \multicolumn{1}{c}{4.15}    & \multicolumn{1}{c}{14.06}     & \multicolumn{1}{c|}{32.42}       & \multicolumn{1}{c}{-}    & \multicolumn{1}{c}{-}     & \multicolumn{1}{c}{-}   \\
        & \multicolumn{1}{c|}{VALUE \cite{li2021value}}       & \multicolumn{1}{c}{5.93}    & \multicolumn{1}{c}{18.76}     & \multicolumn{1}{c|}{-}        & \multicolumn{1}{c}{3.01}    & \multicolumn{1}{c}{7.80}     & \multicolumn{1}{c}{-}       \\
        & \multicolumn{1}{c|}{Ours}       & \multicolumn{1}{c}{\textbf{7.57}}    & \multicolumn{1}{c}{\textbf{21.20}}     & \multicolumn{1}{c|}{-}       & \multicolumn{1}{c}{\textbf{4.01}}    & \multicolumn{1}{c}{8.11}     & \multicolumn{1}{c}{-}        \\
        \hline
        \hline
        \multicolumn{1}{c|}{\multirow{3}{*}{\rotatebox[origin=c]{90}{TEST}}} & \multicolumn{1}{c|}{HERO \cite{li2020hero}}       & \multicolumn{1}{c}{6.21}    & \multicolumn{1}{c}{19.34}     & \multicolumn{1}{c|}{36.66}      & \multicolumn{1}{c}{-}    & \multicolumn{1}{c}{-}     & \multicolumn{1}{c}{-}           \\
         & \multicolumn{1}{c|}{VALUE \cite{li2021value}}       & \multicolumn{1}{c}{6.39}    & \multicolumn{1}{c}{19.54}     & \multicolumn{1}{c|}{-}        & \multicolumn{1}{c}{1.90}    & \multicolumn{1}{c}{6.17}     & \multicolumn{1}{c}{-}       \\
        & \multicolumn{1}{c|}{Ours}       & \multicolumn{1}{c}{\textbf{8.17}}    & \multicolumn{1}{c}{\textbf{21.95}}     & \multicolumn{1}{c|}{-}       & \multicolumn{1}{c}{\textbf{3.64}}    & \multicolumn{1}{c}{6.17}     & \multicolumn{1}{c}{-}        \\ \hline
    \end{tabular}
    }
    \end{center}
    \caption{results on VCMR datasets}
    \label{tab:vcmr results}
\end{table}

The results of VCMR task on TVR and How2R datasets are reported on Table \ref{tab:vcmr results}. 
Besides the baseline model HERO \cite{li2020hero} and the value benchmark \cite{li2021value}, we compare our model with XML \cite{lei2020tvr}, HAMMER \cite{zhang2020hierarchical} and ReLoCLNet \cite{zhang2021video}. 
Our method outperforms all the other methods on TVR dataset, and has slight advantage over Recall@1 metric on How2R dataset.
Specifically speaking, to some extent, we render that XML, ReLoCLNet and HERO use the same 
kind of approach called late fusion, since video retrieval and moment retrieval are all based on similarities between videos and query texts. 
HAMMER and our method are considered as the same method which performs two stage inference and adopt fine-grained cross-modal integration. 
This result demonstrates that fusing video with query promotes the performance in video moment retrieval. 
Compared to HAMMER, we use multi-channel inputs, this explains why our result surpasses its. 
Note that although HERO conducted experiments on How2R dataset, the version used by HERO is not the same as the one used by VALUE benchmark and us, so we cannot extract their results on How2R.

\subsection{Performance Comparison on VR}

\begin{table}[htbp]
    \footnotesize
    \begin{center}
    \setlength{\tabcolsep}{1.1mm}{
    \begin{tabular}{c|c|cccc|cccc}
         \hline
         
         \multicolumn{1}{c|}{}&\multicolumn{1}{c|}{\multirow{2}{*}{Method}} & \multicolumn{4}{c|}{YC2R} &\multicolumn{4}{c}{VATEX-EN-R}\\
         
         & & {\scriptsize R@1} & {\scriptsize R@5} & {\scriptsize R@10} & {\scriptsize AveR} & {\scriptsize R@1} & {\scriptsize R@5} & {\scriptsize R@10} & {\scriptsize AveR}\\
         \hline
         \hline
        \multicolumn{1}{c|}{\multirow{2}{*}{\rotatebox[origin=c]{90}{VAL}}} & \multicolumn{1}{c|}{VALUE}       & \multicolumn{1}{c}{26.23}    & \multicolumn{1}{c}{49.71}     & \multicolumn{1}{c}{60.05}    & \multicolumn{1}{c|}{45.33}    & \multicolumn{1}{c}{55.62}    & \multicolumn{1}{c}{89.15}     & \multicolumn{1}{c}{95.07}   & \multicolumn{1}{c}{79.95}     \\
        & \multicolumn{1}{c|}{Ours}       & \multicolumn{1}{c}{\textbf{32.05}}    & \multicolumn{1}{c}{\textbf{55.47}}     & \multicolumn{1}{c}{\textbf{65.01}}   & \multicolumn{1}{c|}{\textbf{50.84}}      & \multicolumn{1}{c}{38.93}    & \multicolumn{1}{c}{76.13}     & \multicolumn{1}{c}{86.09}     & \multicolumn{1}{c}{67.05}   \\
        \hline
        \hline
        \multicolumn{1}{c|}{\multirow{2}{*}{\rotatebox[origin=c]{90}{TEST}}} & \multicolumn{1}{c|}{VALUE}       & \multicolumn{1}{c}{35.10}    & \multicolumn{1}{c}{59.48}     & \multicolumn{1}{c}{68.27}   & \multicolumn{1}{c|}{54.28}     & \multicolumn{1}{c}{24.51}    & \multicolumn{1}{c}{54.34}     & \multicolumn{1}{c}{68.41}     & \multicolumn{1}{c}{49.09}  \\
        & \multicolumn{1}{c|}{Ours}       & \multicolumn{1}{c}{\textbf{41.21}}    & \multicolumn{1}{c}{\textbf{65.90}}     & \multicolumn{1}{c}{\textbf{75.56}}   & \multicolumn{1}{c|}{\textbf{60.89}}      & \multicolumn{1}{c}{24.04}    & \multicolumn{1}{c}{54.15}     & \multicolumn{1}{c}{68.62}   & \multicolumn{1}{c}{48.94}       \\ \hline
    \end{tabular}
    }
    \end{center}
    \caption{results on VR datasets}
    \label{tab:vr results}
\end{table}

Table \ref{tab:vr results} shows the model performances on two VR datasets. 
By experiment we find that CLIP-ViT+S3D does not work on YC2R, therefore we use ResNet+S3D feature on YC2R dataset. 
The large gap between VALUE and ours on validation set and slight performace difference on test set demonstrate that model is easy to overfit on VATEX-EN-R dataset, as well as mentioned in \cite{li2021value}.

\subsection{Ablation Study}

\begin{table}[htbp]
    \footnotesize
    \begin{center}
    \setlength{\tabcolsep}{1.3mm}{
    \begin{tabular}{c|cccc|cccc}
         \hline
         
         \multicolumn{1}{c|}{\multirow{2}{*}{Method}} & \multicolumn{4}{c|}{TVR} &\multicolumn{4}{c}{How2R}\\
         
         & {\scriptsize R@1} & {\scriptsize R@5} & {\scriptsize R@10} & {\scriptsize AveR} & {\scriptsize R@1} & {\scriptsize R@5} & {\scriptsize R@10} & {\scriptsize AveR}\\
         \hline
         \hline
        \multicolumn{1}{c|}{Base}       & \multicolumn{1}{c}{6.66}    & \multicolumn{1}{c}{14.78}     & \multicolumn{1}{c}{19.39}    & \multicolumn{1}{c|}{13.61}    & \multicolumn{1}{c}{2.54}    & \multicolumn{1}{c}{4.71}     & \multicolumn{1}{c}{6.56}   & \multicolumn{1}{c}{4.61}    \\
        \multicolumn{1}{c|}{B+Fusion}       & \multicolumn{1}{c}{7.34}    & \multicolumn{1}{c}{15.37}     & \multicolumn{1}{c}{20.37}   & \multicolumn{1}{c|}{14.36}    & \multicolumn{1}{c}{\textbf{4.01}}    & \multicolumn{1}{c}{5.48}     & \multicolumn{1}{c}{7.18}     & \multicolumn{1}{c}{5.56}   \\
        \multicolumn{1}{c|}{B+F+Aug}       & \multicolumn{1}{c}{\textbf{7.81}}    & \multicolumn{1}{c}{15.69}     & \multicolumn{1}{c}{20.49}   & \multicolumn{1}{c|}{14.67}    & \multicolumn{1}{c}{3.47}    & \multicolumn{1}{c}{5.94}     & \multicolumn{1}{c}{7.57}     & \multicolumn{1}{c}{5.66}   \\
        \multicolumn{1}{c|}{B+F+A+PT}       & \multicolumn{1}{c}{7.57}    & \multicolumn{1}{c}{\textbf{16.21}}     & \multicolumn{1}{c}{\textbf{21.20}}    & \multicolumn{1}{c|}{\textbf{14.96}}   & \multicolumn{1}{c}{\textbf{4.01}}    & \multicolumn{1}{c}{\textbf{6.56}}     & \multicolumn{1}{c}{\textbf{8.11}}      & \multicolumn{1}{c}{\textbf{6.23}}  \\
    \end{tabular}
    }
    \end{center}
    \caption{Ablation results}
    \label{tab:ablation}
\end{table}

Since our fusion method focus on VCMR task, we only study ablations on two VCMR datasets. 
The results show that the cross-modal integration improves VCMR performance by a large margin. It proves that fusing video with query can absorb more information leading to more precise video moment localization result. 
And data augmentions including shuffling, cutoff and dropout contribute to overall metrics, despite of decreasing Recall@1 on How2R dataset.
Pre-training on four datasets also promotes the performance on VCMR task.

\section{Conclusion}

Although VCMR task is extention of SVMR task, some commonly used methods in SVMR are not employed in VCMR.
In this paper, we add a video-query feature fusion approach to video retrieval to further align features and fuse them at a fine-grained level.
This approach significantly improves the performance of the model on VCMR task. 
Meanwhile, data augmentation methods such as token shuffling, cutoff and dropout effectively improve the robustness of the model and further enhance the model performance.

{\small
\bibliographystyle{ieee_fullname}
\bibliography{egbib}
}

\end{document}